\documentclass{article}

\usepackage{graphicx}
\graphicspath{ {output/plots} }
\usepackage{float}
\usepackage{hyperref}       
\usepackage{caption}
\usepackage{subcaption}
\usepackage{booktabs}
\usepackage{authblk}
\usepackage{lipsum}
\usepackage[numbers]{natbib}
\usepackage{amsmath}
\usepackage{xcolor}
\usepackage{rotating}

\title{Forecasting mortality associated emergency department crowding}

\def\correspondingauthor{\footnote{
	Corrersponding author.

	\emph{Email address:} jalmari.nevanlinna@tuni.fi
}}

\author[1]{Jalmari Nevanlinna\correspondingauthor{}}
\author[1,2]{Anna Eidstø}
\author[1,2]{Jari Ylä-Mattila}
\author[4]{Teemu Koivistoinen}
\author[1,3]{Niku Oksala}
\author[5]{Juho Kanniainen}
\author[1,4]{Ari Palomäki}
\author[1]{Antti Roine}

\affil[1]{\footnotesize Faculty of Medicine and Health Technology, Tampere University}
\affil[2]{\footnotesize Emergency Department, Tampere University Hospital}
\affil[3]{\footnotesize Centre for Vascular Surgery and Interventional Radiology, Tampere University Hospital}
\affil[4]{\footnotesize Kanta-Häme Central Hospital, Hämeenlinna, Finland}
\affil[5]{\footnotesize Faculty of Information Technology and Communication Sciences, Tampere University}

\begin{document}
\maketitle

\AtBeginEnvironment{tabular}{\small}
\AtBeginEnvironment{caption}{\normal}

\begin{abstract}
    Emergency department (ED) crowding is a global public health issue that has been repeatedly associated with increased mortality. Predicting future service demand would enable preventative measures aiming to eliminate crowding along with it's detrimental effects. Recent findings in our ED indicate that occupancy ratios exceeding 90\% are associated with increased 10-day mortality. In this paper, we aim to predict these crisis periods using retrospective data from a large Nordic ED with a LightGBM model. We provide predictions for the whole ED and individually for it's different operational sections. We demonstrate that afternoon crowding can be predicted at 11 a.m. with an AUC of 0.82 (95\% CI 0.78-0.86) and at 8 a.m. with an AUC up to 0.79 (95\% CI 0.75-0.83). Consequently we show that forecasting mortality-associated crowding using anonymous administrative data is feasible.
\end{abstract}

\section{Introduction}\label{introduction}
Emergency department crowding remains a persistent problem worldwide, the adverse effects of which include increased number of medication errors \cite{Kulstad2010}, delays in onset of medication such as antibiotics \cite{Pines2007}, analgesics \cite{Pines2008} or thrombolysis \cite{Schull2003}, prolonged length of stay \cite{McCarthy2009} and dissatisfaction of the patients \cite{Boudreaux2000, Sun2000}. Most importantly, however, crowding has been repeatedly associated with increased mortality \cite{Richardson2006, Guttmann2011, Zhang2019, Sun2013, Ugglas2021, Jo2014, Eidsto2023, Jones2022}. There are multiple temporal and local reasons for a single crowding event, but the recurrent, global and worsening nature of the phenomenon point to a more systemic underlying cause: aging populations and difficulty to hire the required medical personnel to satisfy the ever increasing demand \cite{Morley2018}. In Europe, as populations have been projected to age up until year 2100 \cite{eurostat}, there is no quick and spontaneous resolution of the problem in sight.

For these reasons, there has been a continued interest in optimizing the utilization of the limited resources that are readily available. One manifestation of this effort is emergency department forecasting that has been of academic interest since 1981 \cite{Diehl1981} with over 100 articles published ever since \cite{Gul2018}. Over the last few years, machine learning has taken over traditional statistical models in the field of time series forecasting \cite{Makridakis2022, Makridakis2020} and this been reflected in ED forecasting as well. For example, we recently demonstrated the superiority of LightGBM over both traditional statistical and other novel machine learning models \cite{Tuominen2023} and the performance of ensemble models has been investigated by others as well \cite{Petsis2022, Alvarez2023}. Some tailored deep learning architectures have also been proposed \cite{Sharafat2021, Harrou2020} while some more traditional statistical learning approaches are still being investigated \cite{Reboredo2023}. Despite these contributions, several important gaps in the literature remain.

First, vast majority of previous studies have focused on predicting patient volumes or occupancy in continuous terms \cite{Gul2018}. This is an obvious and intuitive approach, but it has two problems: the resulting metrics are difficult to communicate to an ED stakeholder and impossible to transfer to other facilities. Continuous error metrics such as the mean absolute error or root mean squared error are useful when models are compared against each other with the same sample. But they are not useful in answering the question that really matters: is this performance accurate enough for deployment in the clinic? As colleagues \citet{Hoot2009} so elegantly put it in 2009:

\begin{quote}
    An early warning system must incorporate two components: 1) a clear definition of a crisis period, and 2) a means of predicting crises.
\end{quote}

For the last 15 years, the first -- and the most important -- part of this wisdom seems to have been forgotten, excluding few notable exceptions \cite{Xie2022}. The definition of a crisis period is important because it is the pre-requisite for action. After all, forecasting is never an end in itself but a means to an end: regardless of the context it always aims to enable pre-emptive maneuvers that prevent an adverse outcome in the future. So what is this adverse outcome or \emph{a clear definition of a crisis period} in the context of an ED? We believe the answer to be self-evident: the ED is in crisis when patient safety becomes compromised due to the crowded state alone. There is increasing evidence that certain emergency department occupancy ratios (EDOR) with increased mortality. Most of them have simply compared the most crowded quartile with the less crowded ones \cite{Richardson2006, Jo2014}, which serves to document the existence of the association, but does not reveal the specific threshold for mortality associated crowding. We recently showed that an EDOR exceeding 90\% is associated with increased 10-day mortality in a large Nordic combined ED \cite{Eidsto2023}. In this study, this threshold serves as the definition of a crisis period and consequently as the target variable of interest.

Second, the majority of previous effort has focused on forecasting aggregated visit statistics such as total arrivals or occupancy. This is a natural first step, especially if the focus of the work is in developing algorithms or designing deep learning architectures \cite{Sharafat2021, Harrou2020, Reboredo2023}. However, aggregated visit statistics can be misleading. For example, the Nordic combined ED under the investigation in this article consists of several largely independent sections with their dedicated personnel. A section can be crowded without directly affecting the others and for example crowding of the medical section without crowding of the surgical section is common. Moreover, the implications of crowding vary significantly depending on the respective section. For example, the crowding of the medical section, where elder patients within more comorbidities are common, is likely far more dangerous than crowding of the walk-in section where most patients present with lower acuity conditions. Thus, it is important to account for this heterogeneity when forecasting the future. Surprisingly, we are aware of only one study by \citet{Aroua2015} in which this was accounted for by using diagnosis-based stratification.

Third, the access to health data has become more stringent because of  the emerging national and European regulation. While this carries benefits for ensuring privacy, extensive regulation has constraining effects for both innovation and research as has been recently demonstrated \cite{peukert2022, bessen2020, Bruck2023}. In fact, the current legislation makes simulation-based approaches -- as proposed e.g. by \citet{Hoot2008} -- difficult if not impossible to implement. For this reason, the models in this study have been designed to operate with anonymous administrative time series data.

In this study, we aim to address these gaps by developing an early warning algorithm for ED crowding that i) predicts mortality associated crowding; ii) performs stratified predictions for operational sections of the ED; iii) does not require access to personal information to operate; and iv) provides sufficient temporal margin for action. 

\section{Materials and Methods}\label{materials_and_methods}
\subsection{Data sets and data splitting} 

Tampere University Hospital is an academic teaching hospital located in Tampere, Finland. Tampere University Hospital serves as a secondary care provider for over 500,000 residents within the Wellbeing Service County of Pirkanmaa and is the sole facility in the region equipped to handle all severe emergencies. Additionally, it functions as a tertiary care unit for a broader area encompassing more than 900,000 residents. With approximately 90,000 annual visits, its ED ranks among the largest in the Nordic European countries. The ED features a total of 65 beds, allocated as follows: 6 beds for resuscitation, 36 for medical treatment, and 23 for surgical treatment. Additionally, there is a waiting area for walk-in patients who do not require continuous monitoring. Critical patients exhibiting significant disturbances in vital functions are treated in the resuscitation room and are given priority over others.

In this study, we use a dataset collected for the purposes of our previous work by \citet{Eidsto2023} which contains all episodes of care among bedoccupying patients in our ED spanning from January 1, 2018 to February 29, 2020. We refer to the original publication for detailed description of the sample and data handling protocols. For the purposes of this study, four emergency department occupancy ratio (EDOR) time series were generated based on the original dataset for i) all bedoccupying, ii) medical, iii) surgical and iv) critical patients. All the time series were generated on hourly resolution and predictions were made on the higher volume time series i-iii whereas iv was only used an explanatory variable.

\paragraph{Re-training protocol and testing}

The whole dataset contained 791 days (18,973 hours) of data. The models were evaluated using data from January 1, 2019 to February 29, 2020 spanning a period of 426 days (10,213 hours). Initial training was done using period from January 1, 2018 to December 31, 2018 spanning 365 days (8760 hours) after of which the model was consecutively retrained for each day of the test set resulting in expanding window cross validation.

\subsubsection{Target variables}

\begin{figure}[H]
    \centering
        \includegraphics[width=0.5\textwidth]{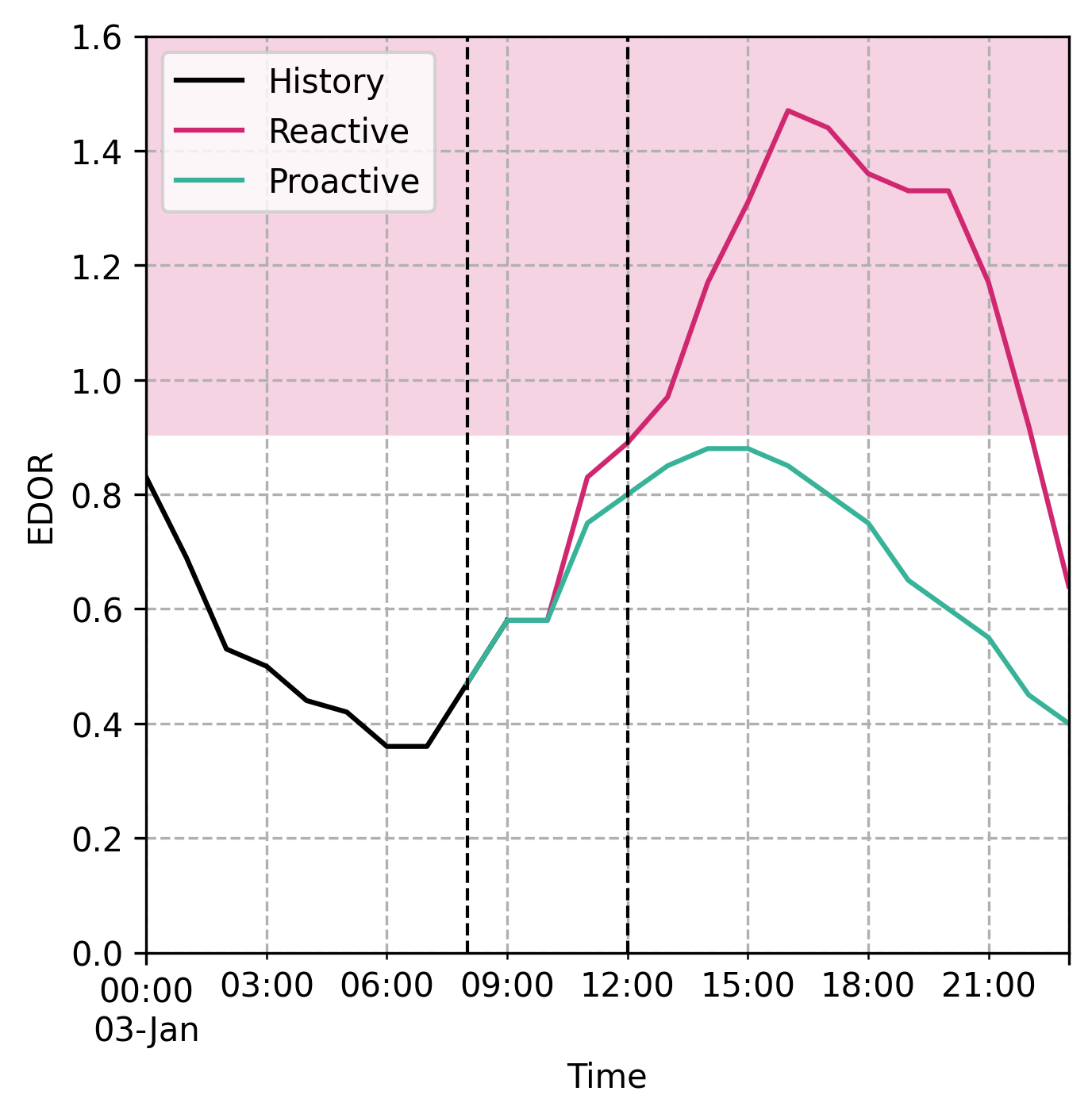}
        \caption{Graphical representation of the concept of the early waning software. The red area shows EDOR$\ge$90 \% which we have previously associated with increased 10-day mortality \cite{Eidsto2023}. The black and red lines represent the historical hourly EDOR among bedoccupying patients during January 3, 2018. The vertical dashed line on the right shows the point when patient safety becomes compromised, after of which the occupancy continued to rise until peaking at almost 150\% at 5 p.m. If crowding could be foreseen e.g. at 8 a.m. (dashed vertical line on the left), it would enable pre-emptive measures that would optimally result in EDOR levels shown in green line, and ultimately improved patient safety.}
        \label{fig:concept}
\end{figure}

In our previous work, we associated EDOR levels higher than 90\% with increased 10-day mortality \cite{Eidsto2023}. Here, we will consider every day during of which three or more hours exceed this threshold to be crowded. In our testing protocol we simulate passing of time during a morning shift from 8 a.m. to 1 p.m. and aim to forecast whether the rest of the day will be crowded or not. This idea is presented graphically in Figure \ref{fig:concept}. Classification results for each forecast origin are reported independently.

\subsubsection{Explanatory variables}

The explanatory variables are listed in Table \ref{tab:explanatory_variables}. For each day, six different forecast origins were considered from 8 a.m. to 1 p.m. and the origins were used as an indicator variable along with subgroup indicator variables. For each origin, both EDOR and crowding history of each subgroup for the last 168 hours were included. Five weather variables were included: precipitation, snow depth and air temperature maximum, minimum and mean. This resulted in $24 \times 1365$ matrix for each day in the dataset. This data handling protocol allows us to train a single model at each re-training iteration which enables faster training and allows cross-learning across different subsections.

\subsection{Feature importance analysis}

Relative importance of the used explanatory variables (features) were calculated using Shapley additive explanations (SHAP) \cite{Lundberg2017}. Shapley importance, derived from cooperative game theory, is a method used to fairly distribute the total gains (or importance) among features based on their contributions. For a given model, the Shapley value for a feature represents its average marginal contribution across all possible subsets of features. Mathematically, for a set of features \( N \) and a feature \( i \), the Shapley value \( \phi_i \) is given by 
\[ 
\phi_i = \sum_{S \subseteq N \setminus \{i\}} \frac{|S|! (|N| - |S| - 1)!}{|N|!} [v(S \cup \{i\}) - v(S)], 
\] 
where \( S \) is a subset of features not including \( i \) and \( v(S) \) is the value (e.g., predictive accuracy) of the model trained on subset \( S \). This method ensures a fair distribution of feature importance, considering all possible combinations and interactions between features.

\subsection{Performance metrics}

The performance of the model is evaluated using an exhaustive set of binary performance metrics. Some of them are shortly defined below and the rest of them a provided in Appendix \ref{appendix}.

The F1 score is the harmonic mean of precision and recall. It provides a balance between the precision and the recall, making it useful for situations where both false positives and false negatives are important.

\begin{equation}
F1 = 2 \cdot \frac{\text{Precision} \cdot \text{Recall}}{\text{Precision} + \text{Recall}}
\end{equation}

Accuracy measures the overall correctness of the model by calculating the proportion of true results (both true positives and true negatives) among the total number of cases examined.

\begin{equation}
\text{Accuracy} = \frac{\text{TP} + \text{TN}}{\text{TP} + \text{TN} + \text{FP} + \text{FN}}
\end{equation}

Area under the receiver operating characteristics curve (AUROC) provides a single metric to evaluate the performance of a binary classifier. It represents the probability that a randomly chosen positive instance is ranked higher than a randomly chosen negative instance.

\begin{equation}
\text{AUROC} = \int_0^1 \text{TPR}(\text{FPR}^{-1}(x)) \, dx
\end{equation}

Area under the precision-recall curve (AUPRC) measures the trade-off between precision and recall for different threshold values, offering a summary of the model’s performance when dealing with imbalanced datasets.

\begin{equation}
\text{AUPRC} = \int_0^1 \text{Precision}(\text{Recall}^{-1}(x)) \, dx
\end{equation}

\paragraph{Statistical significance}

95\% confidence intervals of the AUROC values were calculated using bootstrapping. Bootstrapping involves drawing a large number \( B \) of bootstrap samples \( X^* = \{x_1^*, x_2^*, \ldots, x_n^*\} \) by sampling with replacement from an original sample \( X = \{x_1, x_2, \ldots, x_n\} \) of size \( n \). For each bootstrap sample \( X^* \), the statistic of interest, \( \theta^* = \theta(X^*) \), is computed. The bootstrap distribution of \( \theta^* \) approximates the sampling distribution of the statistic \( \theta \), allowing for the estimation of parameters such as the mean, variance, and confidence intervals of \( \theta \). In this study n=200.

\subsection{Model}

LightGBM (Light Gradient Boosting Machine) is a machine learning framework for gradient boosting developed \citet{Ke2017}. It is designed to be highly efficient and scalable, particularly for handling large datasets and high-dimensional data. LightGBM is the state-of-the art model in time series in general \cite{Makridakis2022} and we have previously demonstrated it's excellent performance with ED as well \cite{Tuominen2023}.

LightGBM uses a histogram-based learning method that buckets continuous feature values into discrete bins, reducing computational cost and memory usage. It grows trees leaf-wise, splitting the leaf with the highest loss reduction, which allows for deeper and more accurate trees compared to traditional level-wise growth. To handle high-dimensional data, LightGBM employs Exclusive Feature Bundling (EFB), which bundles mutually exclusive features to reduce the number of effective features. This results in faster training without sacrificing accuracy. Additionally, Gradient-based One-Side Sampling (GOSS) focuses on instances with large gradient values, enhancing both training speed and model accuracy by prioritizing harder-to-fit instances. It also directly handles categorical features using a specialized algorithm for decision splitting, avoiding the need for one-hot encoding. This improves efficiency and simplifies the preprocessing pipeline.

\begin{table}[h!]
    \centering
    \caption{Explanatory variables}
    \begin{tabular}{rl}
        \hline
        \textbf{Number} & \textbf{Variable} \\
        \hline
        1-168 & EDOR history (Medical) \\
        169-336 & EDOR history (Bedoccupying) \\
        337-504 & EDOR history (Surgical) \\
        505-672 & EDOR history (Critical) \\
        673-679 & Calendar variables \\
        678-845 & Crowding history (Medical) \\
        846-1013 & Crowding history (Bedoccupying) \\
        1014-1181 & Crowding history (Surgical) \\
        1182-1349 & Crowding history (Critical) \\
        1350-1357 & Weekly lags \\
        1358-1363 & Weather \\
        1363 & Subgroup \\
        1364 & Crowding \\
        1365 & Origin \\
        \hline
        \end{tabular}
    \label{tab:explanatory_variables}
\end{table}

\section{Results}\label{results}
\subsection{Descriptive statistics}

\begin{figure}[p]
    \centering
        \begin{subfigure}[b]{1\textwidth}
            \includegraphics[width=\textwidth]{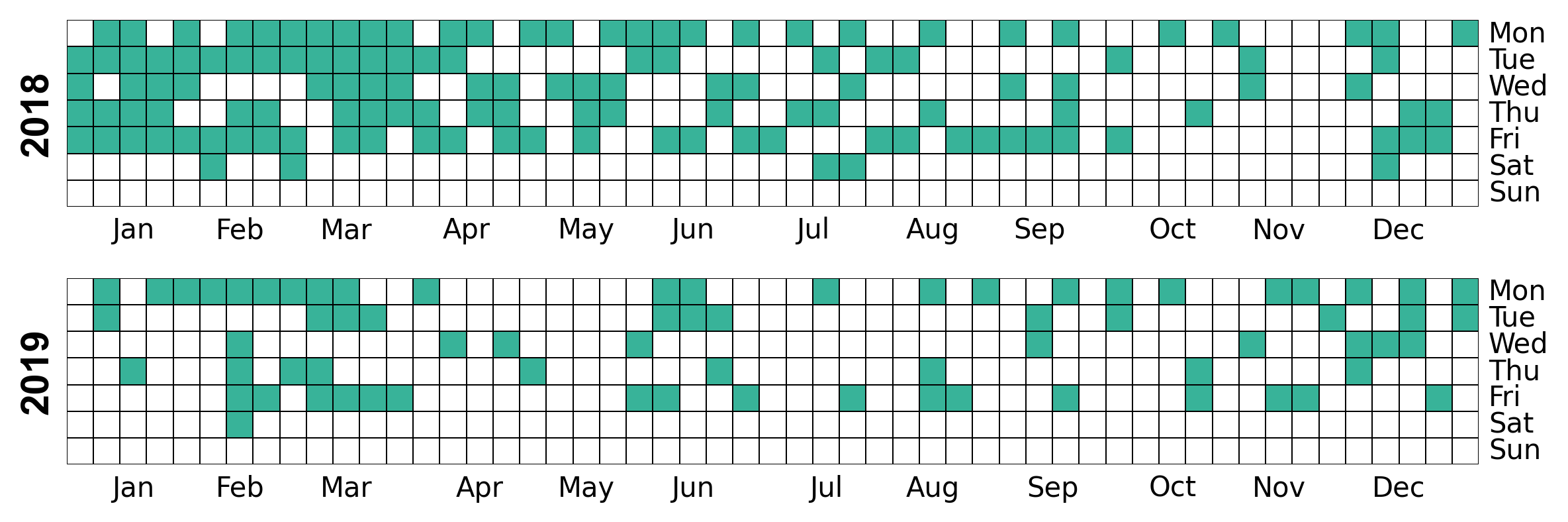}
            \caption{Bedoccupying}
        \end{subfigure}
        \begin{subfigure}[b]{1\textwidth}
            \includegraphics[width=\textwidth]{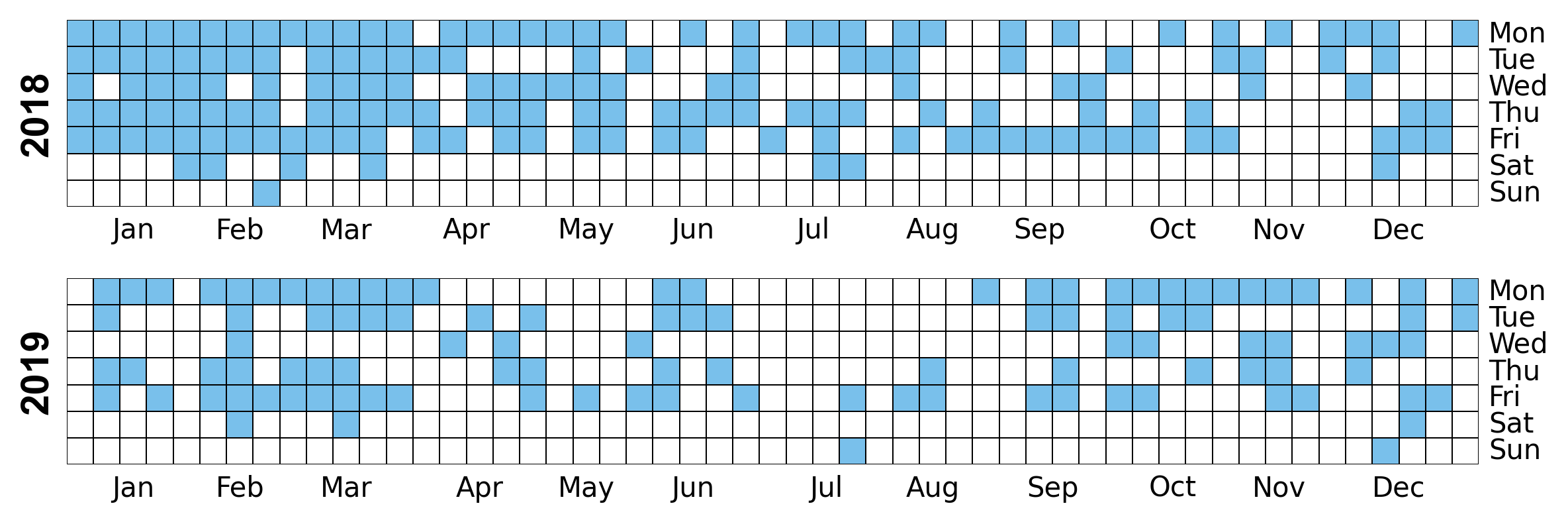}
            \caption{Medical}
        \end{subfigure}
        \begin{subfigure}[b]{1\textwidth}
            \includegraphics[width=\textwidth]{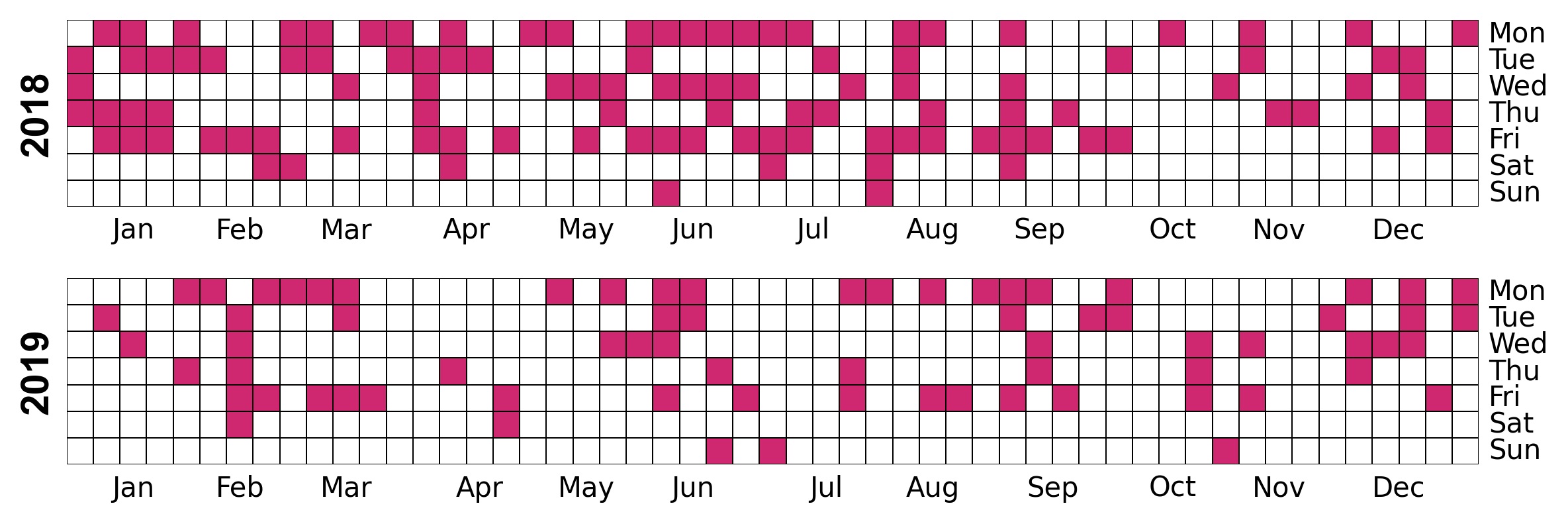}
            \caption{Surgical}
        \end{subfigure}
        \caption{Temporal distribution of crowded days among bedoccupying, medical and surgical patients in the sample. The two months of year 2020 are ommitted here for brevity. The figure demonstrates sporadic occurence of the crowding events that do not follow a deterministic weekday pattern, excluding the relative rarity of crowding during the weekends.}
        \label{fig:calmap}
\end{figure}

\begin{figure}[H]
    \centering  
    \includegraphics[width=1.0\textwidth]{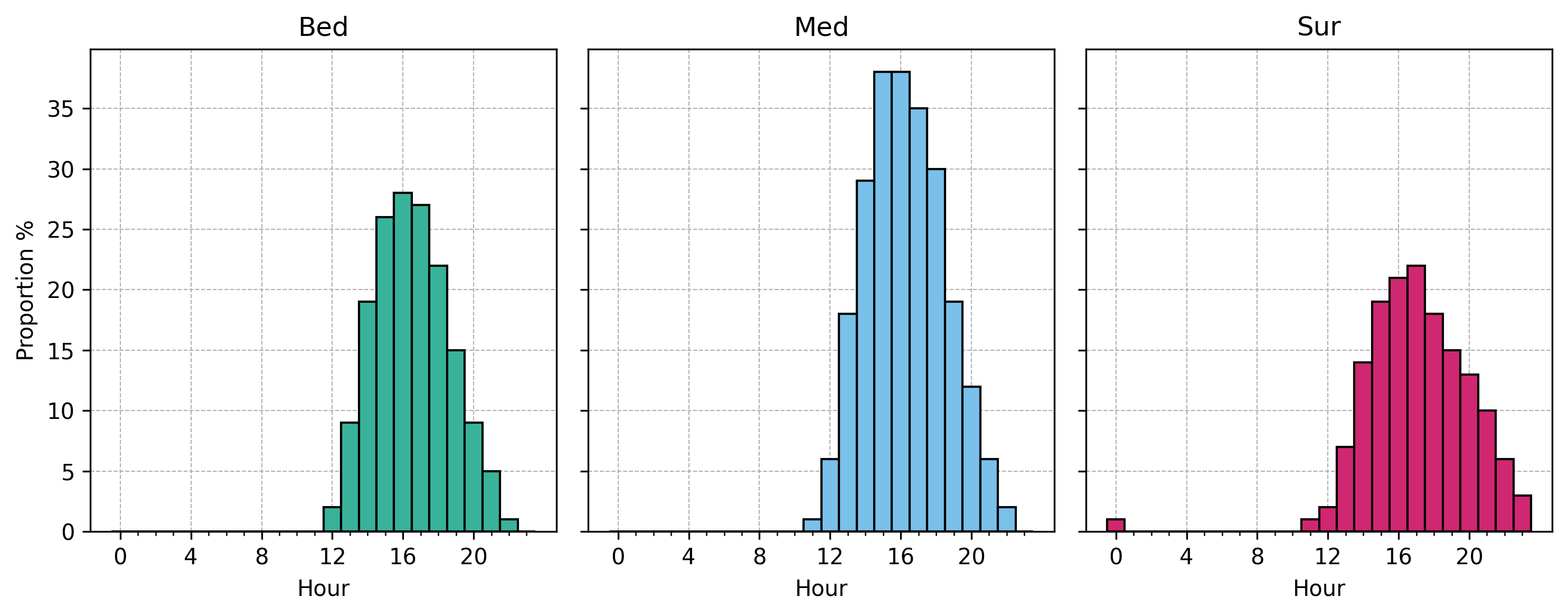}
    \caption{Distribution of crowding events over different hours of the day among bedoccupying, medical and surgical patterns. Note That crowding is nonexistent between 8 a.m. to 11 a.m.}
    \label{fig:bars}
\end{figure}

\begin{figure}[H]
    \centering
        \begin{subfigure}[b]{0.30\textwidth}
            \includegraphics[width=\textwidth]{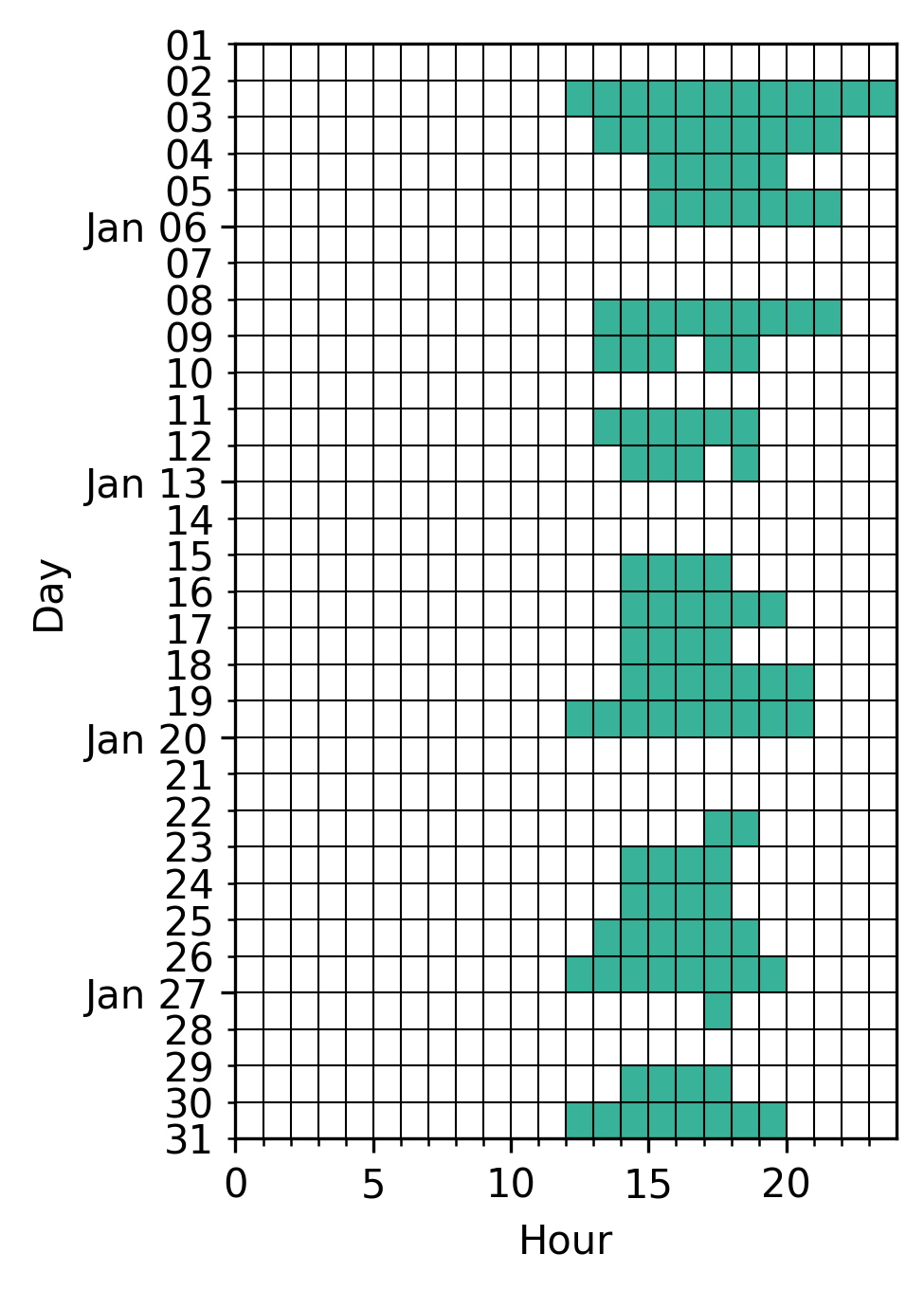}
            \caption{Bedoccupying}
        \end{subfigure}
        \begin{subfigure}[b]{0.30\textwidth}
            \includegraphics[width=\textwidth]{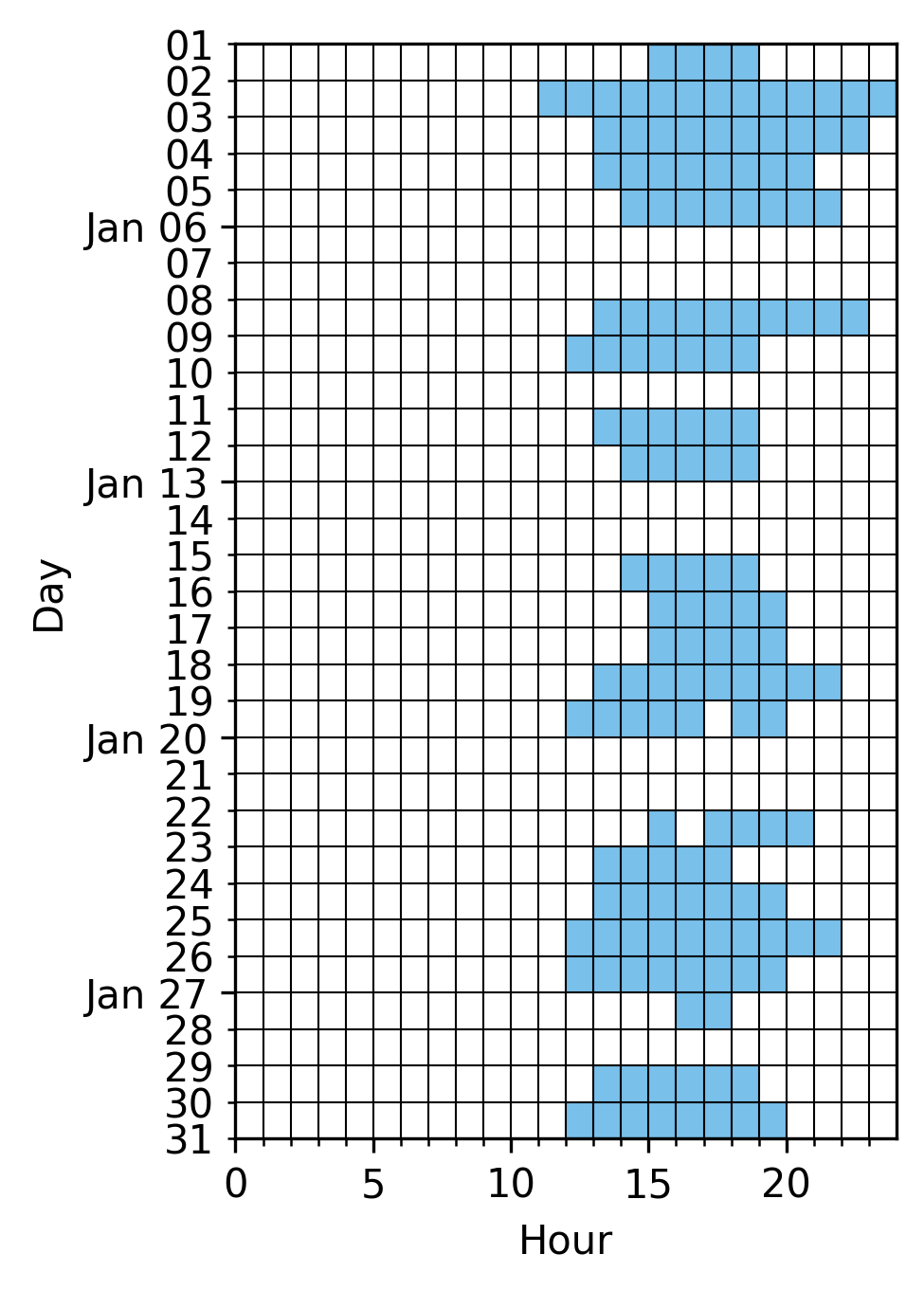}
            \caption{Medical}
        \end{subfigure}
        \begin{subfigure}[b]{0.30\textwidth}
            \includegraphics[width=\textwidth]{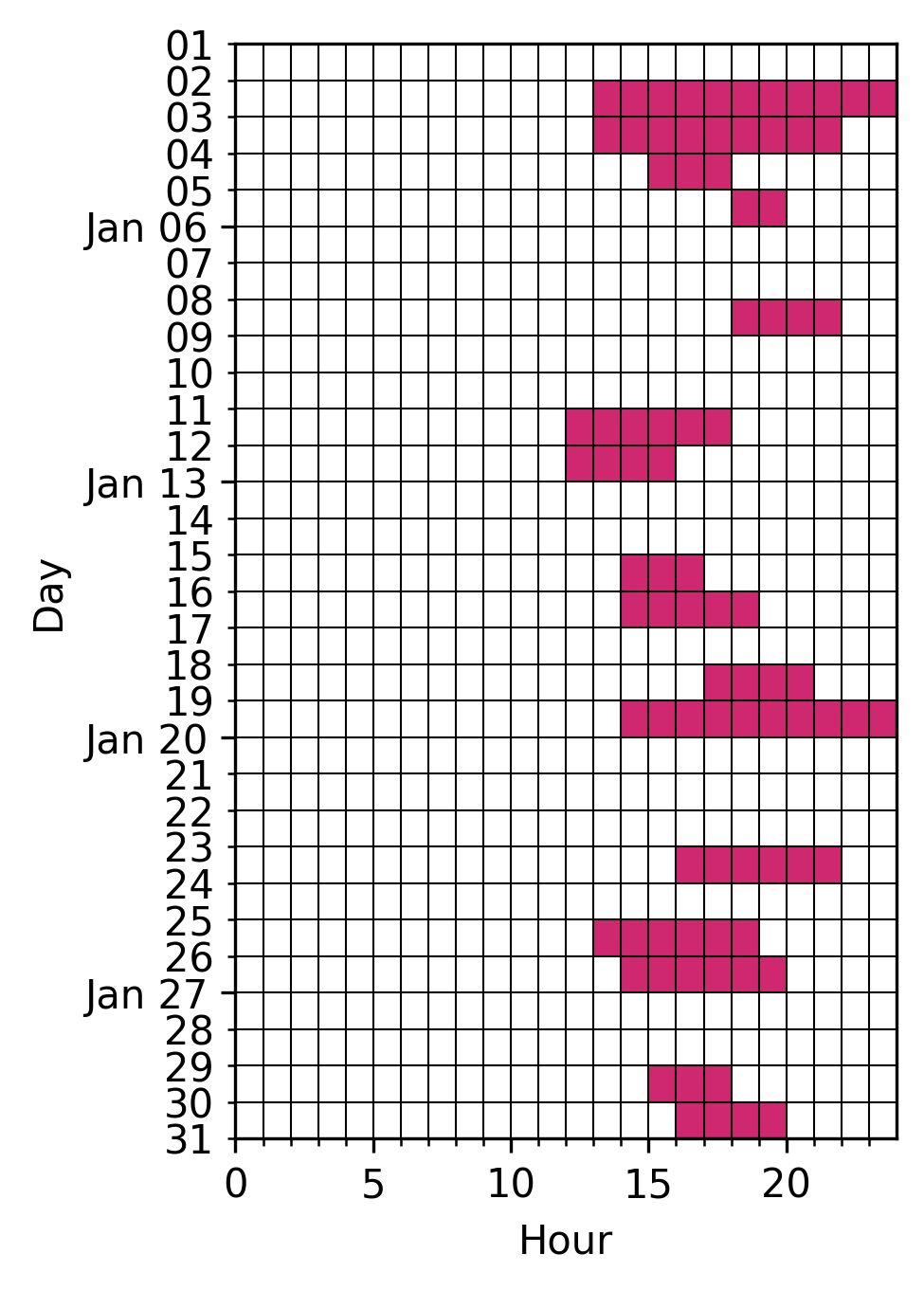}
            \caption{Surgical}
        \end{subfigure}
        \caption{Distribution of the crowding events during January 2018 as a function of hour of the day. Note the varying time of onset and duration of crowding as well as the relative rarity of crowding during the weekends.}
        \label{fig:heatmap}
\end{figure}

Out of the 791 days in the sample, 218 (28 \%) days were crowded among the bedocuppying patients, 288 (36 \%) in the medical section and 199 (25 \%) in the surgical section. Temporal distribution of these days is provided in Figure \ref{fig:calmap}. The hourly pattern of crowding incidence is provided in Figures \ref{fig:bars} and an example on their distribution over several weeks in Figure \ref{fig:heatmap}. Between 8 a.m. and 11 a.m. crowding was nonexistent or extremely rare 0-1 \% of hours being crowded. Crowding starts to increase with prevalence of 2\%, 6\% and 2\% among bedoccupying, medical and surgical sections. After this the prevalence increases following a normal distribution and peaks at 4 p.m. among bedoccupying patients (28\%), 3 p.m. among medical patients (38\%) and 5 p.m. among surgical patients (22\%) before becoming increasingly rare towards the end of the day.

\subsection{Model performance}

\begin{figure}[H]
    \centering
    \begin{subfigure}[b]{0.35\textwidth}
        \includegraphics[width=\textwidth]{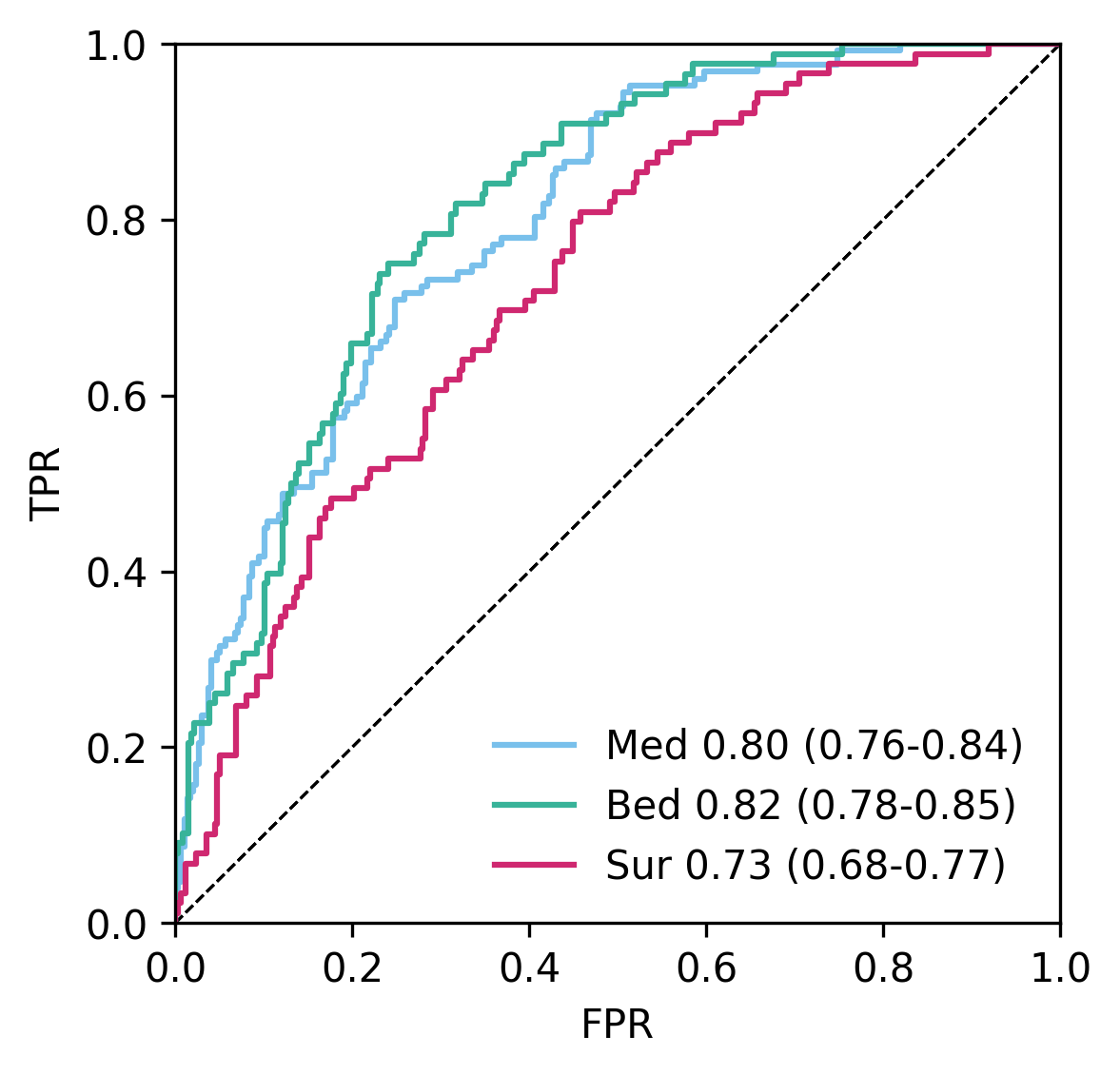}
        \caption{AUROC}
    \end{subfigure}
    \begin{subfigure}[b]{0.57\textwidth}
        \includegraphics[width=\textwidth]{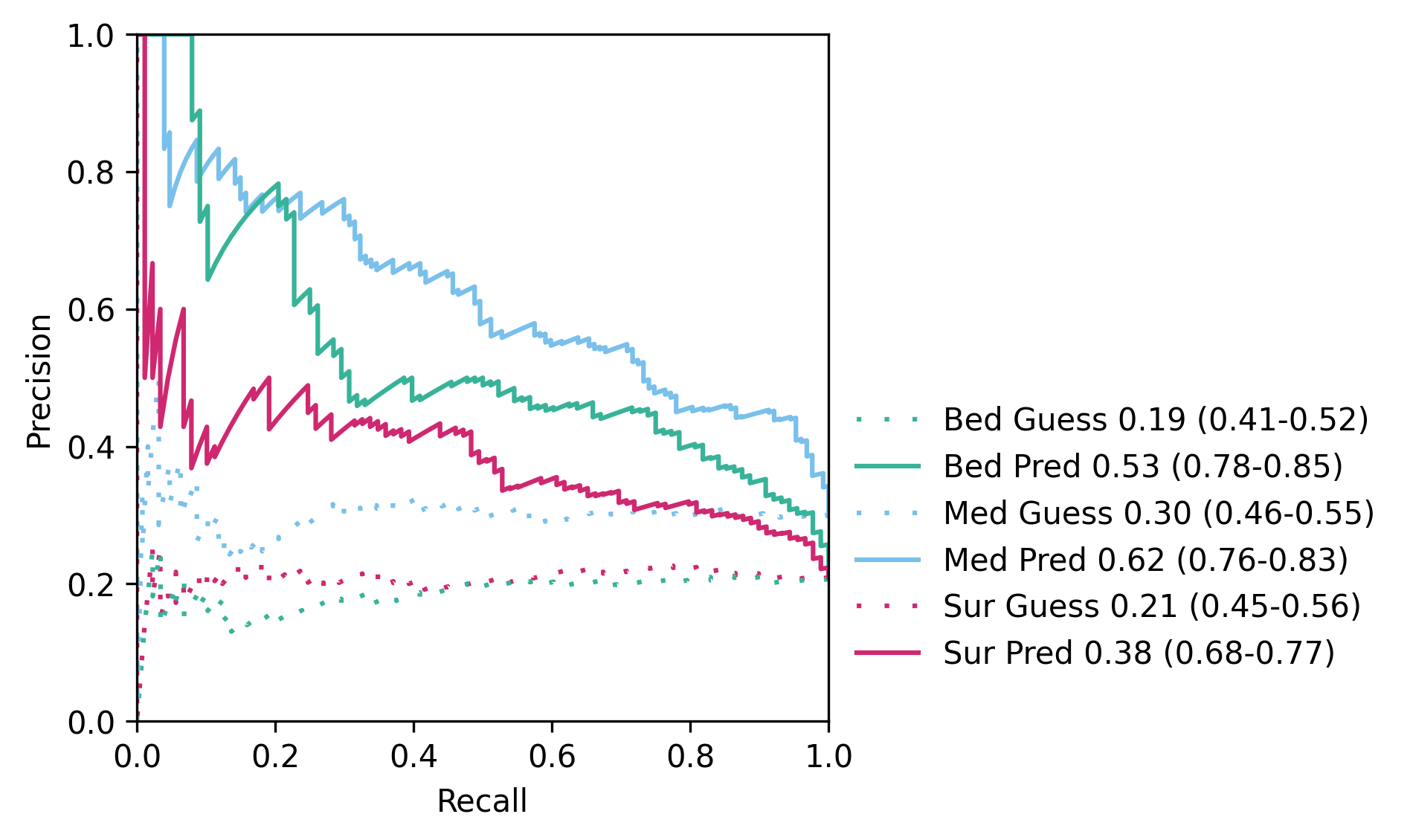}
        \caption{AUPRC}
    \end{subfigure}
    \caption{Area under the receiver operating characteristics curve (AUROC) and area under the precision-recall curve (AUPRC) at forecast origin 11 a.m. Guess level is provided in AUPRC plot for reference. 95\% confidence intervals in parenthesis.}
    \label{fig:auroc_and_auprc}
\end{figure}

Performance metrics of the model among different sections and forecast origins are provided in Table \ref{tab:metrics}. Additionally, AUROC and PRAUC figures at 11 a.m. are provided in Figure \ref{fig:auroc_and_auprc}. Calendar map of the model performance at 11 a.m. among different days on the test set is provided in Figure \ref{fig:calmap}. At 8 a.m. the respective AUC values among bedoccupying, medical and surgical patients were 0.79 (95\% CI 0.75-0.83), 0.77 (95\% CI 0.73-0.80) and 0.74 (95\% CI 0.69-0.78) respectively. The discriminatory ability increased throughout the day so that at 11 a.m. the model reached an AUC of 0.82 (95\% CI 0.78-0.85) among both bedoccupying, 0.80 (95\% CI 0.76-0.84) among medical patients, and 0.73 (95\% CI 0.68-0.77) among surgical patients. At 1 p.m. the respective AUC values were 0.87 (95\% CI 0.84-0.90), 0.86 (95\% 0.83-0.89) and 0.78 (95\% 0.74-0.83) for bedoccupying, medical and surgical patients respectively.

\begin{sidewaystable}
\begin{table}[H]
\caption{
    Binary performance metrics of the model at different forecast origins and sections. 
    TPR = true positive rate, TNR = true negative rate, 
    PPV = positive predictive value,
    NPV = negative predictive value,
    FPR = false positive rate,
    FNR = false negative rate,
    ACC = accuracy,
    AUROC = area under the receiver operating characteristics.
    AUROC 95\% confidence intervals in parenthesis.}
\label{tab:metrics}
\begin{tabular}{lrrrrrrrrrll}
\toprule
Target & Origin & F1 & TPR & TNR & PPV & NPV & FPR & FNR & ACC & AUROC & PRAUC \\
\midrule
Bedoccupying & 8 & 0.47 & 0.58 & 0.77 & 0.40 & 0.88 & 0.23 & 0.42 & 0.73 & 0.79 (0.75-0.83) & 0.44 (0.76-0.83) \\
Bedoccupying & 9 & 0.46 & 0.55 & 0.78 & 0.40 & 0.87 & 0.22 & 0.45 & 0.73 & 0.78 (0.74-0.82) & 0.44 (0.74-0.83) \\
Bedoccupying & 10 & 0.43 & 0.50 & 0.78 & 0.38 & 0.86 & 0.22 & 0.50 & 0.72 & 0.79 (0.75-0.83) & 0.46 (0.75-0.82) \\
Bedoccupying & 11 & 0.52 & 0.57 & 0.83 & 0.47 & 0.88 & 0.17 & 0.43 & 0.78 & 0.82 (0.77-0.85) & 0.53 (0.77-0.85) \\
Bedoccupying & 12 & 0.56 & 0.62 & 0.84 & 0.50 & 0.90 & 0.16 & 0.38 & 0.80 & 0.84 (0.80-0.88) & 0.60 (0.81-0.88) \\
Bedoccupying & 13 & 0.61 & 0.67 & 0.86 & 0.55 & 0.91 & 0.14 & 0.33 & 0.82 & 0.87 (0.84-0.91) & 0.68 (0.84-0.90) \\
Medical & 8 & 0.57 & 0.71 & 0.67 & 0.48 & 0.84 & 0.33 & 0.29 & 0.68 & 0.77 (0.73-0.81) & 0.56 (0.73-0.81) \\
Medical & 9 & 0.58 & 0.73 & 0.66 & 0.48 & 0.85 & 0.34 & 0.27 & 0.68 & 0.77 (0.73-0.81) & 0.54 (0.73-0.81) \\
Medical & 10 & 0.57 & 0.70 & 0.67 & 0.48 & 0.84 & 0.33 & 0.30 & 0.68 & 0.77 (0.73-0.80) & 0.53 (0.73-0.81) \\
Medical & 11 & 0.61 & 0.72 & 0.71 & 0.52 & 0.86 & 0.29 & 0.28 & 0.72 & 0.80 (0.76-0.84) & 0.62 (0.77-0.83) \\
Medical & 12 & 0.63 & 0.72 & 0.76 & 0.56 & 0.86 & 0.24 & 0.28 & 0.75 & 0.83 (0.80-0.86) & 0.68 (0.79-0.86) \\
Medical & 13 & 0.65 & 0.73 & 0.78 & 0.59 & 0.87 & 0.22 & 0.27 & 0.77 & 0.86 (0.83-0.89) & 0.75 (0.83-0.89) \\
Surgical & 8 & 0.40 & 0.42 & 0.82 & 0.38 & 0.84 & 0.18 & 0.58 & 0.73 & 0.74 (0.69-0.78) & 0.36 (0.69-0.78) \\
Surgical & 9 & 0.42 & 0.47 & 0.80 & 0.39 & 0.85 & 0.20 & 0.53 & 0.73 & 0.74 (0.69-0.78) & 0.39 (0.69-0.77) \\
Surgical & 10 & 0.40 & 0.44 & 0.80 & 0.37 & 0.84 & 0.20 & 0.56 & 0.73 & 0.72 (0.66-0.77) & 0.35 (0.68-0.76) \\
Surgical & 11 & 0.43 & 0.44 & 0.84 & 0.42 & 0.85 & 0.16 & 0.56 & 0.76 & 0.73 (0.68-0.77) & 0.38 (0.68-0.77) \\
Surgical & 12 & 0.47 & 0.47 & 0.86 & 0.47 & 0.86 & 0.14 & 0.53 & 0.78 & 0.76 (0.72-0.80) & 0.45 (0.71-0.80) \\
Surgical & 13 & 0.50 & 0.51 & 0.87 & 0.50 & 0.87 & 0.13 & 0.49 & 0.79 & 0.78 (0.74-0.82) & 0.52 (0.73-0.83) \\
\bottomrule
\end{tabular}
\end{table}

\end{sidewaystable}

\begin{figure}[H]
    \centering
        \begin{subfigure}[b]{1.0\textwidth}
            \includegraphics[width=\textwidth]{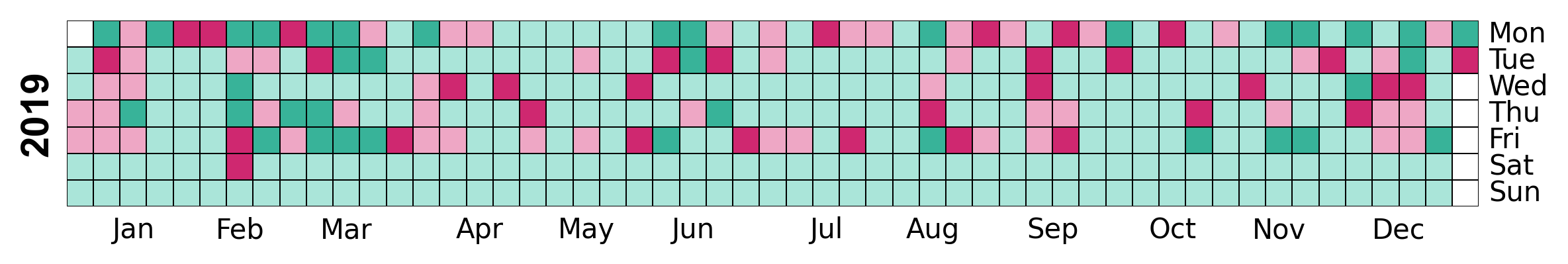}
            \caption{Bedoccupying}
        \end{subfigure}
        \begin{subfigure}[b]{1.0\textwidth}
            \includegraphics[width=\textwidth]{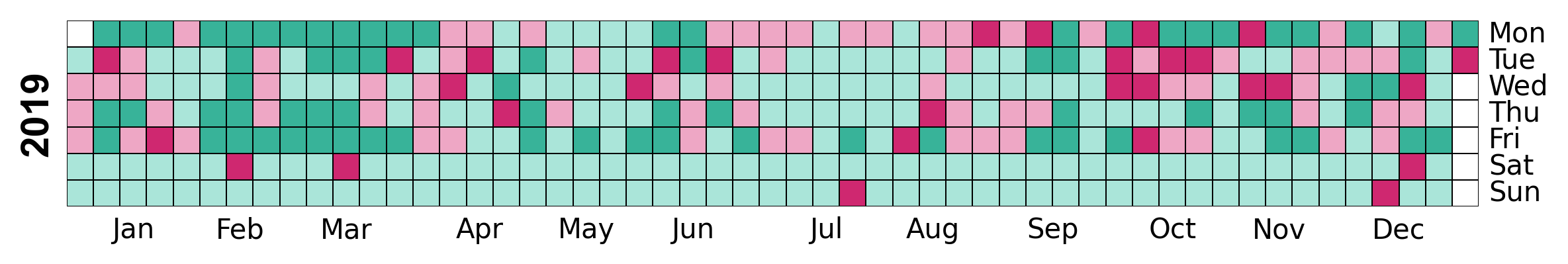}
            \caption{Medical}
        \end{subfigure}
        \begin{subfigure}[b]{1.0\textwidth}
            \includegraphics[width=\textwidth]{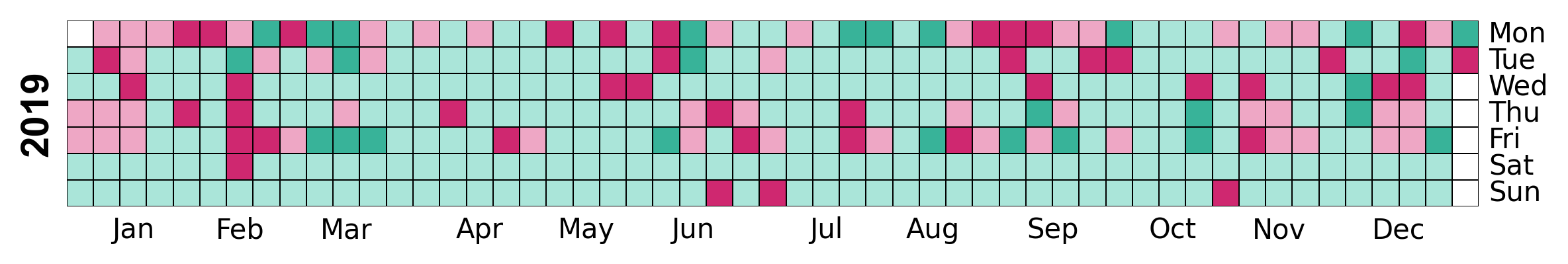}
            \caption{Surgical}
        \end{subfigure}
        \caption{Performance at origin 11. Dark green: true positive, Dark red: false positive, Light green: true negative, Light red: false negative}
\end{figure}

\paragraph{Feature importance} 

\begin{figure}[H]
    \centering
        \includegraphics[width=\textwidth]{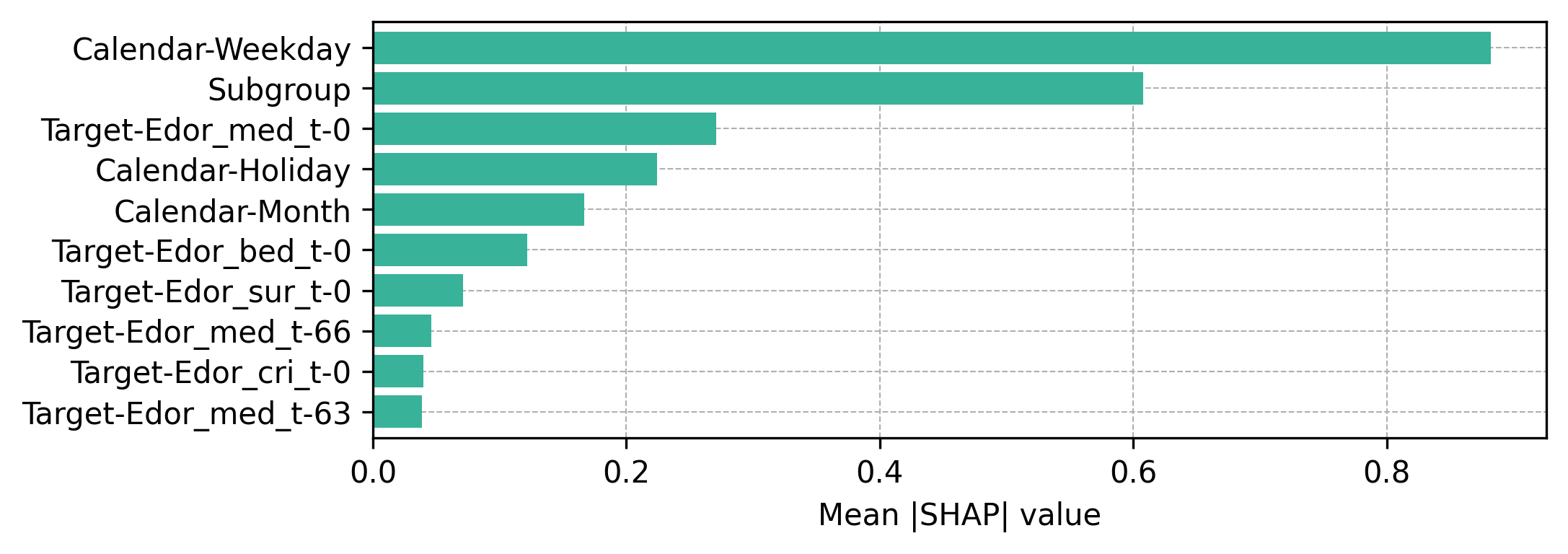}
        \caption{Feature importance statistics as estimated using mean absolute Shapley values}
        \label{fig:importance}
\end{figure}

Feature importance statistics are provided in Figure \ref{fig:importance}. Weekday was the most important feature, followed by subgroup and current EDOR of the medical section. Holiday was the fourth most important feature.

\section{Discussion}\label{discussion}
This study had three main findings. Each of them are discussed separately below.

\paragraph{Forecasting is possible} First, we showed that forecasting mortality associated crowding using very simple administrative data is possible with sufficient temporal margin for action. The discriminatory power of the model reached an excellent level in terms of AUC at 11 a.m. and demonstrated fair performance from 8 a.m. to 11 a.m. In fact, the model matches or exceeds the performance of many clinicial decision support algorithms that are widely used in everyday clinical practice. For example, National early warning score (NEWS) that is widely used to predict in-hospital mortality has been documented to reach an AUC of 0.73 \cite{Eckart2019}. At the default threshold, the negative predictive value of the model was good at the expense of positive predictive value.


\paragraph{Sections differ from one another} The model performed better in predicting future crowding among medical and bedoccuping patients compared to surgical patients. The underlying cause remains elusive with the current dataset. We hypothesize that lower prevalence of crowding among surgical section might play a role. Additionally, medical patients include a sizeable segment of frail patients for whose ability to cope at home is compromised by relatively small disturbances. The presentation of these patients at the ED is likely more correlated with the state of primary care, which may lead to more predictable patterns.

\paragraph{Calendar variables are important} \texttt{Weekday} was the most important feature which is not surprising because the weekly seasonality of occupancy is well known based on clinical experience and repeatedly documented academically. Coincidentally, \citet{Petsis2022} recently also used Shapley values in their work and the day of week emerged as the most important feature. This is because weekends are typically relatively quiet whereas Mondays and Fridays tend to be the most crowded days due to the pent up demand caused by the weekend. This phenomenon can actually be seen to some extent in Figures \ref{fig:calmap} and \ref{fig:heatmap} as well. The importance of the calendar variables was also reflected in high feature importance value of the \texttt{holiday} indicator variable. The relative importance of the current occupancy of the medical section is logical because developing crowding in the medical section can precede crowding of the whole ED or the surgical section. 


\subsection{Limitations}

There were some limitations in this study. First, the definition of crowding does not account for different durations of the event although the implications can differ significantly between three hours and eight hours of consecutive crowding. Second, both training and testings set were relatively small which might result in understating the performance of the model but these kind of problems are inevitable if the system would be implemented in practice. Third, it is important to remember that forecasts alone do not help anyone. In order to extract their benefits, they have to be coupled with effective interventions, the most important of which is ensuring the availability of follow-up care beds as recently highlighted by \citet{Stewart2024}. Fourth, this study was limited by design to work with anonymous administrative data and avoided using personal health data to comply with current regulation. It is possible that the performance can be further improved by incorporating more nuanced information about the status of the ED.

\section{Conclusions}\label{conclusions}
In this study, we were set out to investigate whether mortality associated ED crowding can be predicted with sufficient margin for action. Our results suggest that 1) forecasting mortality-associated crowding is feasible using anonymous administrative data, 2) LightGBM model demonstrates high predictive accuracy, particularly for medical and bedoccupying patients and achieves an AUC of 0.80 by 11 a.m, 3) sections differ from one another in terms of both prevalence of crowding and predictability and 4) predicting is possible without access to patient level data which makes the model implementable regardless of the current privacy regulation. The study highlights the need for integrating these forecasts with actionable interventions to enhance patient safety and optimize ED operations. Future research should explore the benefit of including more granular data and involve clinical stakeholders in moving the model from \emph{in silico} to \emph{in vivo}.

\bibliographystyle{plainnat}
\bibliography{references}

\appendix
\newpage
\section{Appendix}\label{appendix}

True positive rate (TPR), also known as recall, measures the proportion of actual positives that are correctly identified by the model.

\begin{equation}
\text{TPR} = \frac{\text{TP}}{\text{TP} + \text{FN}}
\end{equation}

True negative rate (TNR), also known as specificity, measures the proportion of actual negatives that are correctly identified by the model.

\begin{equation}
\text{TNR} = \frac{\text{TN}}{\text{TN} + \text{FP}}
\end{equation}

Positive predictive value (PPV), also known as precision, measures the proportion of positive results that are true positives.

\begin{equation}
\text{PPV} = \frac{\text{TP}}{\text{TP} + \text{FP}}
\end{equation}

Negative predictive value (NPV) measures the proportion of negative results that are true negatives: 

\begin{equation}
\text{NPV} = \frac{\text{TN}}{\text{TN} + \text{FN}}
\end{equation}

False positive rate (FPR) measures the proportion of actual negatives that are incorrectly identified as positives by the model:

\begin{equation}
\text{FPR} = \frac{\text{FP}}{\text{FP} + \text{TN}}
\end{equation}

False negative rate (FNR) measures the proportion of actual positives that are incorrectly identified as negatives by the model.

\begin{equation}
    \text{FNR} = \frac{\text{FN}}{\text{FN} + \text{TP}}
\end{equation}

\end{document}